\documentclass[10pt,twocolumn,letterpaper]{article}

\usepackage[pagenumbers]{cvpr} 

\usepackage{graphicx}
\usepackage{amsmath}
\usepackage{amssymb}
\usepackage{booktabs}
\usepackage{xcolor,colortbl}
\usepackage{float}
\usepackage{balance}
\usepackage[pagebackref,breaklinks,colorlinks]{hyperref}

\usepackage[capitalize]{cleveref}
\crefname{section}{Sec.}{Secs.}
\Crefname{section}{Section}{Sections}
\Crefname{table}{Table}{Tables}
\crefname{table}{Tab.}{Tabs.}

\def\cvprPaperID{2457} 
\def\confName{CVPR}
\def\confYear{2023}

\begin{document}

\title{Progressive Text-to-Image Generation}
\author{Zhengcong Fei, Mingyuan Fan, Li Zhu, Junshi Huang\thanks{The corresponding author.} \\
Meituan\\
Beijing, China\\
{\tt\small \{name\}@meituan.com}}

\twocolumn[{%
\maketitle
\vspace*{-1.2cm}
\begin{figure}[H]
\hsize=\textwidth 
\centering
\includegraphics[width=2\columnwidth]{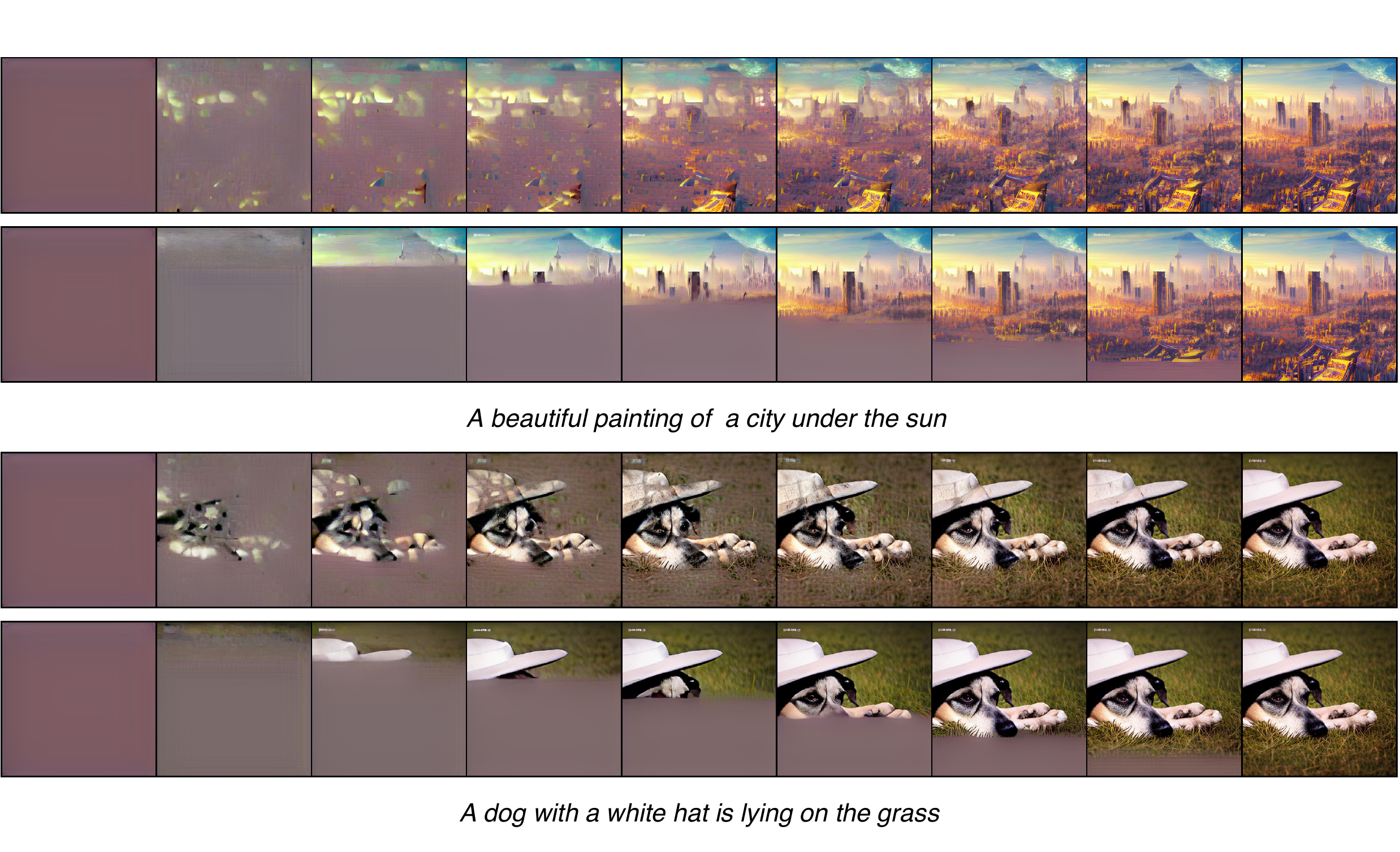} 
\caption{Illustration of different generation orders for text-to-image synthesis. Conventional model generates vector quantized image sequence \emph{from left to right} as top, while our progressive model creates image patches \emph{from coarse to fine} as bottom. }
\label{fig:1}
\end{figure}
}]

\begin{abstract}
Recently, Vector Quantized AutoRegressive (VQ-AR) models have shown remarkable results in text-to-image synthesis by equally predicting discrete image tokens from the top left to bottom right in the latent space. Although the simple generative process surprisingly works well, is this the best way to generate the image? For instance, human creation is more inclined to the outline-to-fine of an image, while VQ-AR models themselves do not consider any relative importance of image patches. In this paper, we present a progressive model for high-fidelity text-to-image generation. The proposed method takes effect by creating new image tokens from coarse to fine based on the existing context in a parallel manner, and this procedure is recursively applied with the proposed error revision mechanism until an image sequence is completed. The resulting coarse-to-fine hierarchy makes the image generation process intuitive and interpretable. Extensive experiments in MS COCO benchmark demonstrate that the progressive model produces significantly better results compared with the previous VQ-AR method in FID score across a wide variety of categories and aspects. Moreover, the design of parallel generation in each step allows more than $\times 13$ inference acceleration with slight performance loss.
\end{abstract}


\section{Introduction}

The task of text-to-image generation aims to create natural and consistent images from the input text and has received extensive research interest. 
Recently, latent autoregressive (AR) generation frameworks have achieved great success in advancing the start-of-the-arts, by learning knowledge and patterns from a large-scale multimodal corpus \cite{radford2021learning,li2021align,yu2022coca}. 
Generally, they treat the task as a form of language modeling and use Transformer-like \cite{vaswani2017attention} architectures to learn the relationship between language inputs and visual outputs. 
A key component of these approaches is the conversion of each image into a sequence of discrete units through the use of a VQ-VAE \cite{van2017neural} based image tokenizer, \emph{e.g.}, VQ-GAN \cite{esser2021taming,yu2021vector}, RQ-VAE \cite{lee2022autoregressive} and ViT VQ-GAN \cite{yu2021vector}.
Visual tokenization essentially unifies the view of text and images so that both can be treated simply as sequences of discrete tokens and is adaptable to sequence-to-sequence models. To that end, DALL-E \cite{ramesh2021zero}, CogView \cite{ding2021cogview}, RQ-Transformer \cite{lee2022autoregressive}, and Parti \cite{yu2022scaling} employ autoregressive models to learn text-to-image task from a large collection of potentially noisy text-image pairs \cite{changpinyo2021conceptual,jia2021scaling, gafni2022make}. In particular, \cite{wu2022nuwa} further expand on this AR over AR modeling approach to support arbitrarily-sized image generation.

Another research line for text-to-image generation involves diffusion-based methods, such as GLIDE \cite{nichol2021glide}, DALL-E 2 \cite{ramesh2022hierarchical}, stable diffusion \cite{rombach2022high}, RQ-Transformer \cite{lee2022autoregressive}, and Imagen \cite{saharia2022photorealistic}.
These models pursue to directly generate images or latent image features with diffusion process \cite{ho2020denoising,dhariwal2021diffusion} and produce high-quality images with great aesthetic appeal. 
Even so, discrete sequence modeling for text-to-image generation remains appealing given extensive prior work on large language models \cite{brown2020language} and advances in discretizing other modalities, such as video and audio, as cross-language tokens \cite{bonddeep}. 
However, the current constructed plain and equal paradigm, without enough global information \cite{tan2021progressive}, may not reflect the progressive hierarchy/granularity from high-level concepts to low-level visual details and is not in line with the actual human image creation. Also, the time complexity of standard auto-regressive image sequence generation is $\mathcal{O}(n)$, which meets a critical limitations for high resolution image generation.

Motivated by the above factors, we present the \emph{progressive model} for text-to-image generation from coarse to fine. 
Specifically, it takes text tokens as inputs to an encoder and progressively predicts discrete image tokens with a decoder in the latent space. The image tokens are then transformed by the VQ-GAN decoder, which can produce high-quality reconstructed outputs. 
As illustrated in Figure \ref{fig:1}, given text prompts, our model first generates high-level content skeleton, then these information are used as pivoting points according to which to create the details of finer granularity. This process iterates until an image is finally completed by adding the fine-grained tokens. Meanwhile, the error tokens generated in previous steps can be \textit{dynamically} revised as more details are filled. 
We show that such progressive generation in a latent space is an effective and efficient way to improve text-to-image performance, enabling to accurately integrate and visually convey world knowledge.

To evaluate the framework, we conduct text-to-image generation experiments on the popular MS COCO \cite{lin2014microsoft} benchmark. Compared with the convention AR model with similar model parameters, our method achieves significantly better image generation performance, as measured by image quality and image-text alignment in both automatic metrics and human evaluations. The progressive model also provides important benefits for the inference speed. As the inference time of AR methods increases linearly with the output image resolution, the progressive model provides the global context for image token prediction and employs the importance score for parallel set selection. This allows us to provide an effective way to achieve a better trade-off between the inference speed and the image quality. We hope this technique can help visual content creators to save time, cut costs and improve their productivity and creativity.

Finally, we summarize the contributions of this paper as follows: 
($\textbf{i}$) \emph{Order matters}. We argue that the importance of image tokens is not equal and present a novel progressive model in the VQ-based latent space for text-to-image generation. Compared with previous work, our method allows long-term control over a generation due to the top-down progressive structure and enjoys a significant reduction over empirical time complexity. ($\textbf{ii}$) We use large-scale pre-training and \emph{dynamic error revision} mechanism customized to our approach, to further boost image generation performance. ($\textbf{iii}$) Experiments on the dataset across different aspects demonstrate the superiority of progressive model over strong baselines. In particular, our approach is simple to understand and implement, yet powerful, and can be leveraged as a building block for future text-to-image synthesis research. 


\section{Background}

\begin{figure*}[t]
\centering
\includegraphics[width=1.95\columnwidth]{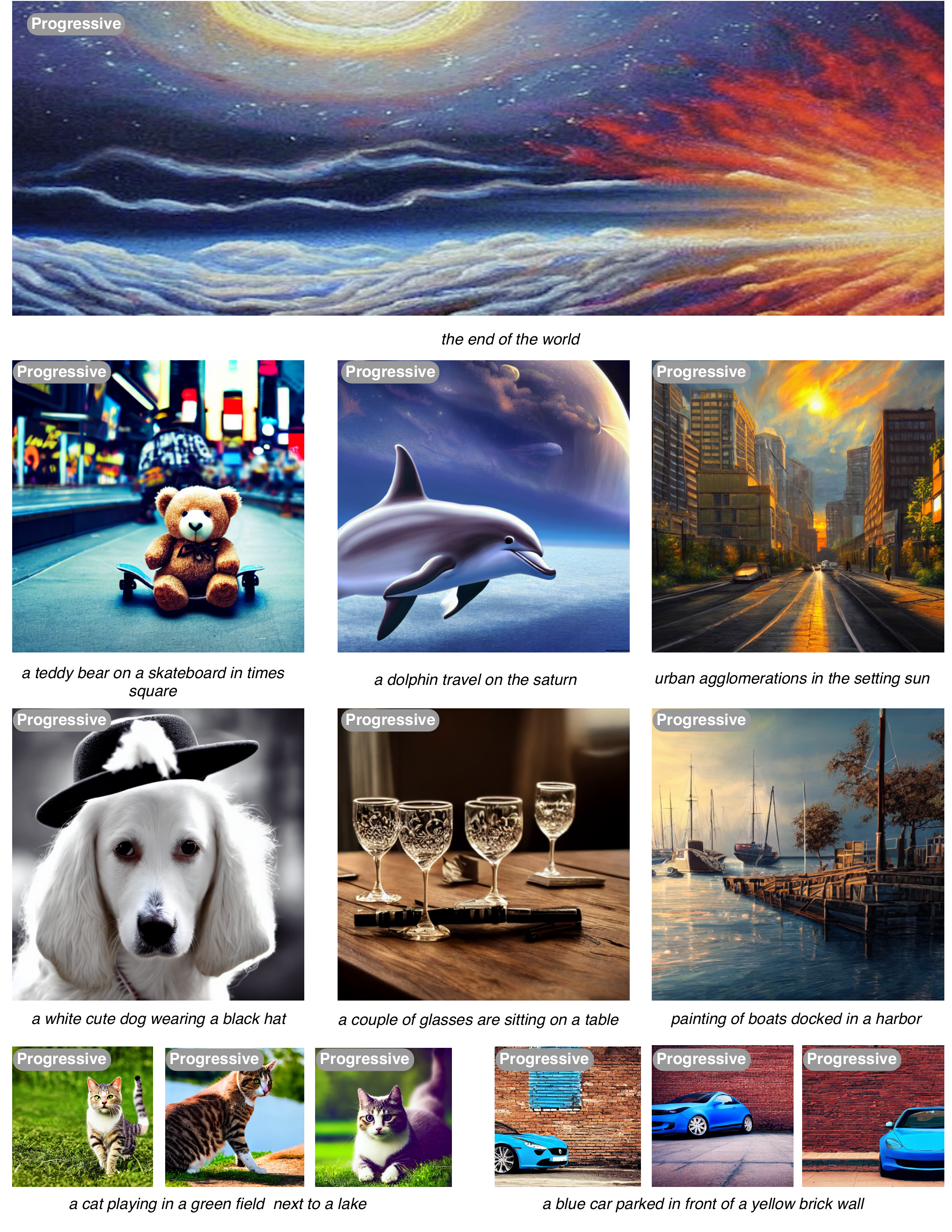} 
\caption{Selected image samples generated from the progressive model and corresponding text prompts. Refer to Sec \ref{sec:4.3} for a more detailed discussion. }
\label{fig:case}
\end{figure*}

We first briefly introduce the conventional vector quantized autoregressive model for text-to-image generation. 
Specifically, the two-stage process includes: ($\textbf{i}$) training an image tokenizer that turns an image into a sequence of discrete visual tokens for training and reconstructs the image at inference time and ($\textbf{ii}$) optimizing a sequence-to-sequence Transformer model that produces image tokens from text tokens in the latent space.

\paragraph{Image Tokenizer.}

Since the computation cost is quadratic to the sequence length, it is limited to directly modeling raw pixels using transformers \cite{chen2020generative}.  
Previous works \cite{van2017neural,yu2021vector} addressed this by using a discrete variational auto-encoder (VAE),  where a visual codebook is learned to map a patch embedding to its nearest codebook entry in the latent space. 
These entries can be considered as visual words, and the appearance of these words of a given image are thus contained image tokens like words in a sentence. 
A VQ-VAE image tokenizer usually follows an encoder-decoder paradigm and is trained with the losses as  \cite{esser2021taming} on the unlabeled images of training data. Specifically, the encoder $E$, the decoder $D$ and the codebook $Z \in \{z_k\}_{k=1}^K$, where $K$ is the code size, can be trained end-to-end via the following loss with training image $I$:
\begin{align} 
\begin{split}
    {L}_{vae} =& ||I - \tilde{I}||_1 + || \text{sg}[E(I) - z_q ] ||_2^2  \\ &+ \beta ||\text{sg}[z_q] - E(I) ||_2^2, 
\end{split}
\end{align}
where $\tilde{I}$ is reconstructed image from $D(z_q)$ and $z_q$ is the indexed embedding from codebook $Z$, that is,
\begin{equation}
    z_q = Q(z) = \text{argmin}_{z_k \in Z} ||z-z_k||_2^2,
\end{equation}
$z = E(I)$ and $Q(\cdot)$ is mapping function from spatial feature $z$ to $z_q$. sg[$\cdot$] stands for the stop-gradient operation. 
In practice, we use VQ-GAN \cite{esser2021taming} with techniques including factorized codes and real-false discriminator for perception loss and exponential moving averages  to update the codebook entries which contribute to training stability and reconstruction quality.

\paragraph{Text-to-Image Transformer.}
After unified image and text modalities with discrete tokens, a standard encoder-decoder Transformer model is then trained by treating text-to-image generation as a sequence-to-sequence modeling problem. The Transformer model takes text prompt as input and is trained using next-token prediction of image latent codes supervised from the image tokenizer. Formally, provided with text prompt $X$, the optimization objective for modeling of image token sequence $Y = \{y_1, \ldots, y_L\}$ in training dataset $\mathcal{D}$ can be factorized as: 
\begin{equation}
	L_{ar} = - \text{log } p(Y|X) = - \text{log } \prod_{i=1}^L p(y_i| y_{<i}, X).
\end{equation}
During inference, the model samples image tokens autoregressively conditioned on the history context, which are later decoded into pixels using the VQ-GAN decoder to create the output image.
For the text encoder, we load a pre-trained BERT-like model \cite{kenton2019bert,raffel2020exploring} for training acceleration, and the decoding part of image tokens is trained from random initialization.
Most of the existing latent-based text-to-image generation models can be split as decoder-only \cite{ramesh2021zero,ding2021cogview}, encoder-decoder \cite{yu2022scaling} and diffusion models \cite{rombach2022high,gu2022vector,wang2022clip,lee2022progressive} in the VQ-VAE based latent space. In this paper, we choose to focus on the encoder-decoder pattern with pre-trained text encoding.

\begin{figure*}[t]
\centering
\includegraphics[width=1.9\columnwidth]{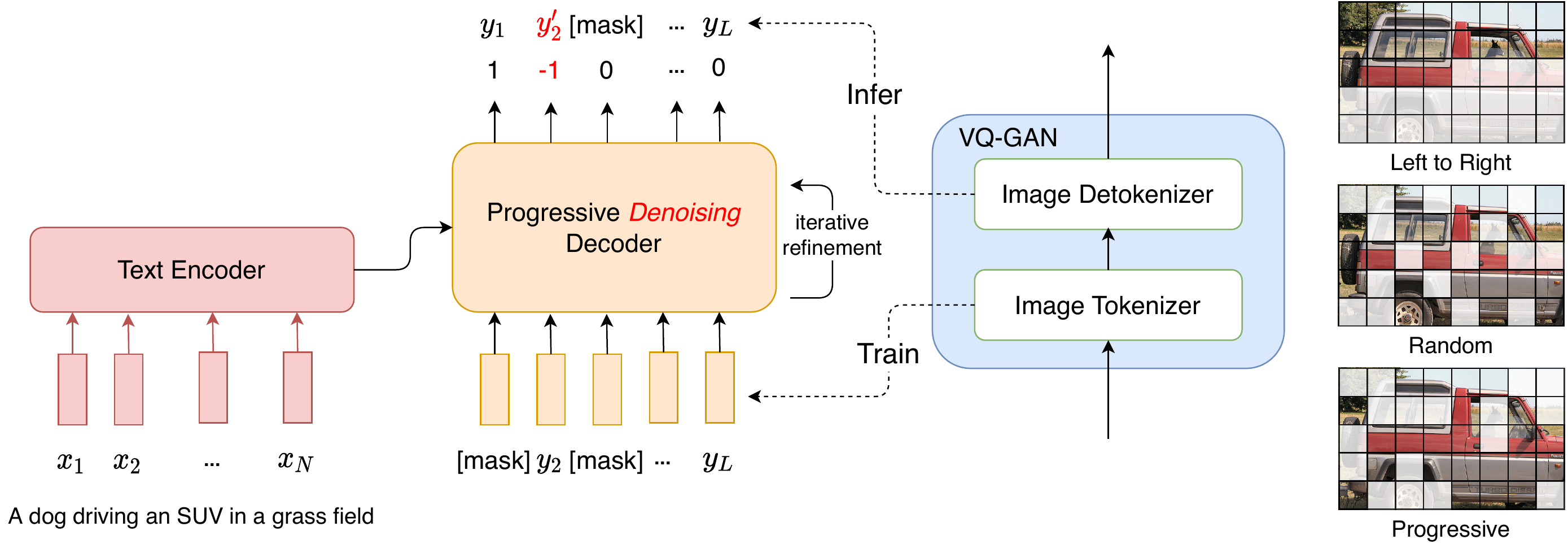} 
\caption{Overview of the proposed progressive text-to-image model, with left-to-right, random, and coarse-to-fine generation orders in the VQ-GAN latent space. Red symbols denote the error revision process.}
\label{fig:3}
\end{figure*}

\section{Methodology}
\label{sec:3}

\subsection{Overview}

Provided with a text prompt $X$, we aim to generate a complete image token sequence $\tilde{Y}$, based on which the reconstructed image from VQ-GAN is accordingly high fidelity and semantically correlative  to $X$. 
The generation procedure of our method can be formulated as a progressive sequence generation of $T$ stages: $\tilde{Y} = \{\tilde{Y}^1, \ldots, \tilde{Y}^T\}$, 
such that the predicted $\tilde{Y}^{t+1}$ at $t+1$-th stage preserves a finer-resolution image sequence compared to the sequence $\tilde{Y}^{\leq t}$ at preceding stages. 
In between, the intermediate state is formed as $\tilde{Y}^t = \{\tilde{y}_1^t, \ldots, \tilde{y}_L^t | \tilde{y}^t_i \in \mathcal{V}\}$ and the corresponding token state sequence $\tilde{Z}^t = \{\tilde{z}_1^t, \ldots, \tilde{z}^t_L | \tilde{z}_i^t \in \{0, 1, -1 \}\}$, where $\mathcal{V}$ is a VQ code vocabulary, and $L$ is the image sequence length. 
The state sequence $\{0, 1, -1\}$ indicate that the token at each position of the intermediate sequence is unchanged, to be generated, or to be replaced, respectively. 
Formally, the $i$-th image token in the intermediate sequence of $(t+1)$-th stage can be further formulated as $(1 - | \tilde{z}_i^{t+1}|) \tilde{y}_i^t + |\tilde{z}_i^{t+1}| \tilde{y}_i^{t+1}$.

\paragraph{Image Sequence Prediction.}
To generate the full sequence of image tokens within $T$ stages, we constraint that $\forall t, \sum_{i=1}^L \mathbb{I}[\tilde{z}_i^t = 1] = \frac{L}{T}$.
Therefore, the generation procedure with $T$ stages can be modeled with factorized conditional probability as: 
\begin{equation}
\small
	p(\tilde{Y}|X) = \prod_{t=1}^T \prod_{i=1}^L p(\tilde{y}_i^{t+1}|\tilde{Y}^{t}, X, \tilde{z}_i^{t+1}) p(\tilde{z}_i^{t+1}| \tilde{Y}^{t}, X, \tilde{Z}^{t}). \label{eq:4}
\end{equation}
At each generation step $t$, model first produces the state sequence $\tilde{Z}^{t+1}$ for the selection of to-be-updated token positions by $p(\tilde{z}_i^{t+1}|\tilde{Y}^{t}, X, \tilde{Z}^{t})$. 
Once the changeable image token positions are determined,  the corresponding tokens are generated or replaced according to distribution $p(\tilde{y}_i^{t+1}|\tilde{Y}^{t}, X, \tilde{Z}^{t+1})$, leading to a new image token sequence $\tilde{Y}^{t+1}$.
Thus, we can recover the final image token sequence $P(\tilde{Y}|X)$ by marginalizing all the intermediate sequences.
Note that such generation procedure starts from a fully masked image sequence $\{[mask], \ldots, [mask]\}$ of length $L$, and then iteratively generate or revise the image tokens according to the predicted state sequence.
Finally, this procedure terminates and outputs final image token sequence $\tilde{Y}^T$ after $T$ steps.

\subsection{Progressive Image Generation}

Two properties are desired for a progressive generation: $i$) important and outline tokens should appear in an earlier stage, so that the generation follows a coarse-to-fine manner; $ii$) the number of stages $T$ should be small enough, thus the generation maintains fast at inference procedure.
To this end, the key points of the progressive model lie in the importance scores of image tokens for the determination of token generation order at each stage, as shown in Figure \ref{fig:3}. In the following paragraphs, we introduce two variants for image token scoring, instead of random selection. 

\paragraph{Quantization Error based Scoring.}

As the quantization error of VQ-GAN reflects the difficulty of image patch discretization, we can decide the generation order of tokens according to the quantization error of image tokens via VQ-GAN.
Intuitively, the smaller the quantization error, the higher the quality and confidence of image to be reconstructed, and thus it is better to be generated at earlier stage.
To this end, we obtain the quantization error sequence $E =\{e_1, \ldots, e_L\}$ for image sequence $Y= \{y_1, \ldots, y_L\}$ with the encoder of VQ-GAN.
At $t$-th step, we collect the index set of top-$\frac{t}{T} L$ smallest values in error sequence $E$, according to which the value of $z^t_i$ in state sequence $Z^t$ is set as 1 if the $i$-th position belongs to the collected index set.
Note that the values of $Z^t$ are initialized as $0$.
Accordingly, the image token sequences $Y=\{Y_1, \ldots, Y_T\}$ can be constructed like the state sequences $Z=\{Z^1, \ldots, Z^T\}$, and we set $Y^0=\{[mask], \ldots, [mask]\}$ of length $L$.

Thereafter, the training instance $(X, Y)$ can be broken into the consecutive series of training tuples $((X, Y^{t-1}), (Z^t, Y^t))$ for $t \in [1, T]$, where $(X, Y^{t-1})$ and $(Z^t, Y^t)$ are the model input and ground-truth label at $t$-th stage, respectively.
Then, we can train the confidence-aware generation model based on $T$ series of training tuples by maximizing $\prod_{i=1}^L p(\tilde{y}_i^{t+1}|Y^{t}, X, z_i^{t+1}) p(\tilde{z}_i^{t+1}| Y^{t}, X, Z^{t})$ at each stage.
In this way, the image tokens with high-confidence are generated at earlier stages, which may serve as the outline of image.
After that, the model can leverage more context information for the generation of uncerntain image tokens at following stages.
As shown in Sec. \ref{sec:model_analysis}, this simple yet powerful scoring strategy learned from pre-defined confidence information presents promising result.

\paragraph{Dynamic Importance Scoring.}

We further propose to learn an dynamic scoring function for individual image token.
In general, the model determines the image token generation order by sampling $\frac{L}{T}$ \textbf{available} positions at each stage, and maximizes a global reward function at final. 
In this case, we refer to the position selection as a policy $\pi_\theta (a_t| s_t)$, where an agent state $s_t$ is $(\tilde{Y}^{\leq t}, X, \tilde{Z}^{\leq t})$.
Note that we consider the entire trajectories of intermediate image sequences and state sequences, \emph{i.e.}, $\tilde{Y}^{\leq t}$ and $\tilde{Z}^{\leq t}$, before and at current $t$-th stage in agent state.
The action $a_t \in \{1, \ldots, L\}^{\frac{L}{T}}$ is an intermediate set, so that the value of $\tilde{Z}^{t+1}_i$ is $1$ if $i \in a_t$, otherwise $0$.
Please note that $a_t$ is sampled without replacement, and $\pi_\theta$ is simply a neural network. 
At the beginning, the agent state $s_1=(\tilde{Y}^0, X, \tilde{Z}^0)$, where $\tilde{Y}^0$ and $\tilde{Z}^0$ are initialized as aforementioned.
Then, we implement the generation process by repeatedly sampling the action $a_t \sim \pi_\theta(s_t)$ and transiting to a new agent state $s_{t+1}$ for $T$ steps. 
Meanwhile, we update the predicted image token sequence $\tilde{Y}^{t+1}$ according to the updated state sequence $\tilde{Z}^{t+1}$ at each $t+1$ stage.
At the final stage, a scalar reward function $r$, \emph{e.g.}, L2 loss for iamge reconstruction \cite{dong2021peco} or CLIP-based similarity \cite{radford2021learning}, is used to find an optimal policy that maximizes the expected global reward.
This procedure is equivalent to minimize the loss function:
\begin{align}
\small
	L_{dis} = -{E}_{\tau \sim \pi_{\theta}(\tau)} [r(s_T)], \\
	\pi_\theta(\tau) = p(s_1)\prod_{t=1}^T p(s_{t+1}|a_t,s_t) \pi_\theta(a_t|s_t),
\end{align}
where $\tau$ is the trajectory $(s_t, a_t, \ldots, s_T, a_T)$, $p(s_1)$ is a deterministic value, and $p(s_{t+1}|a_t,s_t)$ is the generation model to update the image token sequence $\tilde{Y}^{t+1}$ according to the state sequence $\tilde{Z}^{t+1}$ decided by action $a_t$.
In practice, we maximize the reward function by estimating its gradient using the policy gradient \cite{sutton1999policy} strategy.

\subsection{Image Token Revision}
\label{sec:3.3}

Although the preceding progressive training makes the model capable of generating image sequences from coarse to fine, the model cannot alleviate the adverse impact of error tokens from earlier stages.
Particularly, the new image tokens at each stage are simultaneously generated based on the prior context in our paradigm, without considering the information of each other.
Such approach suffers from a conditional independence problem like non-autoregressive generation \cite{gu2018non,gu2021fully}. 
Therefore, it is prone to generate repeated or inconsistent tokens at each generation stage.

To alleviate this problem, we propose an error revision strategy by \emph{injecting pseudo error tokens} into the training data, and helps the model to recover from error tokens generated in previous stages.
Formally, given the training tuple $((X, Y^{t-1}), (Y^t, Z^t))$,  we randomly replace part of image tokens in $Y^{t-1}$ with the tokens from other images, except for $[mask]$ token. 
Meanwhile, the values at corresponding positions of state sequence $Z^t$ are set as -1, which means to-be-updated.
To avoid the misleading caused by too many pseudo error tokens, we randomly select some training tuples of each instance by $\text{Bernoulli}(p_{error})$ for pseudo data generation.
In this way, we construct a new training data $\mathcal{D}$ with those re-built pseudo tuples. 


\begin{table*}[t]
	\begin{center}
	\setlength{\tabcolsep}{2mm}{
		\begin{tabular}{lcccccc}
		& &&&\multicolumn{2}{c}{\textbf{MS COCO FID }($\downarrow$)} \\
		\textbf{Approach}&\textbf{Model Type}&\textbf{\# Data}&\textbf{\# Param}&Zero-shot&Fine-tuned\\
		\hline
		Retrieval Baseline& - &- &-&19.12 &- \\
		\hline
		X-LXMERT&Autoregressive&180K&228M&37.4&-\\
		DALL-E small&Autoregressive&15M&120M&45.8&-\\
		minDALL-E&Autoregressive&15M&1.3B&24.6&-\\
		CogView&Autoregressive &30M&4B&27.1 &-\\
		CogView2&Autoregressive &30M&9B& 24.0&17.7\\
		RQ-Transformer &Autoregressive&30M&3.9B&16.9&-\\
		DALL-E 2 &Diffusion &650M&5.5B&10.39&-\\
		Imagen& Diffusion&860M&7.6B&7.27&-\\
		Parti&Autoregressive&800M&20B&7.23&3.22\\
		 \hline
		PTIG&Progressive&20M&1.2B&13.28&5.34\\
			\hline
		\end{tabular}
		{\caption{FID score comparison of different text-to-image synthesis models on the MS COCO dataset. Some listed evaluation results are from DALL-Eval and corresponding papers. }
			\label{tab:1}}
			}
	\end{center}
\end{table*}

\section{Experiments}

In this section, we conduct the experiments to validate the effectiveness of our progressive model. 
Specifically, we compare with some state-of-the-art approaches in Sec \ref{sec:4.1}, and conduct some ablation studies in Sec \ref{sec:model_analysis}.
Finally, we present some text-to-image cases in Sec \ref{sec:4.3}, and human evaluation in Sec \ref{sec:4.4}.  

The architecture of progressive model almost follows VQ-GAN\cite{esser2021taming} and standard encoder-decoder Transformer paradigm \cite{yu2022scaling}. 
Some slight changes are made: ($\textbf{i}$) remove the causal mask in the image decoder; ($\textbf{ii}$) append the intermediate sequence prediction layers at end of image decoder. 
We use the MS COCO \cite{lin2014microsoft} dataset identical to DALL-Eval \cite{cho2022dall} for performance evaluation in all experiments. For more implementation details please refer to appendix \ref{app:exp_setting}.

Like prior works \cite{yu2022scaling,cho2022dall}, we evaluate the text-to-image generation performance in two primary aspects: generated image quality, and alignment between generated image and input text. Specifically, the evaluation procedures are: 
\begin{itemize}
	\item \emph{Image Quality.}  Fr$\acute{\text{e}}$chet Inception Distance (FID) \cite{heusel2017gans} is used as the primary automated metric for measuring image quality. Concretely, the FID score is computed by inputting generated and real images into the Inception v3 \cite{szegedy2016rethinking} network, and using the output of the last pooling layer as extracted features. The features of the generated and real images are then used to fit two multi-variate Gaussians, respectively. Finally, the FID score is computed by measuring the Fr$\acute{\text{e}}$chet distance between these multi-variate Gaussian distributions.  
	\item \emph{Image-Text Alignment.} Text-image relation degree is estimated by automated captioning evaluation: an image output by the model is captioned with a standard trained Transformer-based model \cite{cornia2020meshed} and then the similarity of the input prompt and the generated caption is assessed via conventional metrics BLEU \cite{papineni2002bleu}, CIDEr \cite{vedantam2015cider}, METEOR \cite{denkowski2014meteor}, and SPICE \cite{anderson2016spice}.
\end{itemize}

\subsection{Comparison with the state-of-the-art methods}
\label{sec:4.1}
In this part, the text-to-image generation performance of different methods is compared. 
Moreover, we introduce a text-to-image retrieval baseline, \emph{i.e.}, given a text prompt, we retrieve the most matched image in the \emph{training set}, measured by the cosine similarity between the text embedding and image embedding from pre-trained CLIP model \cite{radford2021learning}. 

\begin{table*}[t]
	\begin{center}
	\setlength{\tabcolsep}{2mm}{
		\begin{tabular}{lcccc}
		\textbf{Approach}&\textbf{BLEU }($\uparrow$)&\textbf{METEOR }($\uparrow$)&\textbf{CIDEr }($\uparrow$)&\textbf{SPICE }($\uparrow$)\\
		\hline
		Ground Truth (upper bound)&32.5&27.5&108.3&20.4\\
		Retrieval Baseline&22.4&21.7&81.5&15.3\\
		ruDALL-E-XL&13.9&16.0&38.7&8.7\\
		minDALL-E&16.6&17.6&48.0&10.5\\ 
		\hline
		PTIG&20.5&19.7&74.8&13.5\\
			\hline
		\end{tabular}
		{\caption{Comparison for image captioning evaluation on the MS COCO test set.}
			\label{tab:2}}
			}
	\end{center}
\end{table*}

\paragraph{Image Quality Evaluation.}
Following \cite{ramesh2021zero,yu2022scaling}, we use 30,000 generated and real image samples from MS COCO 2014 dataset for evaluation. 
Those images use the same input preprocessing with 256$\times$256 image resolution. 
We compare our proposed models with several state-of-the-art methods, including autoregressive-based models X-LXMERT \cite{cho2020x}, minDALL-E \cite{ramesh2021zero}, 
CogView \cite{ding2021cogview}, CogView 2 \cite{ding2022cogview2}, RQ-Transformer \cite{lee2022autoregressive}, Parti \cite{yu2022scaling}, and diffusion-based models, DALL-E 2 \cite{ramesh2022hierarchical} and Imagen \cite{saharia2022photorealistic}. 
The evaluation results coupled with the size of training data and model parameters are presented in Table \ref{tab:1}. 
We can observe that our progressive model, which has similar parameter size to previous autoregressive-based models, achieves strongly competitive performance while posing an advance in inference speed. In particular, the progressive model shows strong generalization without fine-tuning on specific domains compared with miniDALL-E. 
Besides for scaling more parameters, the experiment results indicate that generation pattern exploration also holds promising potential for text-to-image creation.

\paragraph{Image-text Alignment Evaluation.} 
The evaluation of image-text alignment complements the FID score for text-to-image generation models.
Table \ref{tab:2} presents results of different models on the image-text alignment measurement.
As expected, the progressive model outperforms other popular autoregressive-based models on this metric, and is close to the performance of retrieval-based baseline, which uses retrieval images for captioning. 
However, it should be noted that the results are biased due to the influence caused by the ability of pre-trained image captioning model \cite{cornia2020meshed}.

\begin{table}[t]
\centering
\setlength{\tabcolsep}{2mm}{
\begin{tabular}{lccc}
\hline
\textbf{Order}& \textbf{FID }($\downarrow$) & \textbf{CIDEr }($\uparrow$)\\ 
\hline
Left to right&19.2&54.3\\ 
Random&20.1&51.2\\
Anti-progressive&22.2&49.5\\
Const. Error&16.1&64.3\\
Progressive&13.3&74.8\\
\hline
\end{tabular}}
\caption{Effects of different generation orders.}
\label{tab:3}
\end{table}

\begin{table}[t]
\centering
\setlength{\tabcolsep}{1.7mm}{
\begin{tabular}{lcccc}
\hline
\textbf{\# Stage}&\textbf{FID }($\downarrow$) &\textbf{CIDEr }($\uparrow$)  &\textbf{SpeedUp}\\ \hline
8&19.8&53.5&97.1$\times$\\
16&15.9&65.1&52.5$\times$\\
64&13.3&74.8&13.2$\times$\\
256&12.7&78.5&3.3$\times$\\
1024&12.5&79.8&1$\times$\\
\hline
\end{tabular} }
\caption{Effects of stage numbers.}  \label{tab:4}
\end{table}

\subsection{Model Analysis}
\label{sec:model_analysis}
\paragraph{The Impact of Generation Order.}
To deeply analyze the effectiveness of generation order in text-to-image generation, we compare four different generation strategies under the same experiment setting. 
All baselines predict 16 image tokens at each stage, except for left-to-right manner which predicts 1 image token each time.  As shown in Table \ref{tab:3}, we notice that the synthesis performance drops when replacing the progressive manner with the random or conventional sequence. This may indicate that predicting image tokens from coarse-to-fine benefits the quality of image generation. 
Furthermore, dynamic scoring-based order shows more advance than quantization error scoring and other baselines. 
Interestingly, we also train the model with anti-progressive order, \emph{i.e.}, training model in a fine-to-coarse manner, and we can observe a significant reduction of performance, affirming the value of coarse-to-fine progressive generation manner again. 

\paragraph{Number of Progressive Stages.} 
One of the motivations of this work is that the generation can be parallel at each stage, leading to a significant reduction in training and inference time. 
To achieve the trade-off between inference speed and generation performance, we evaluate the models with different progressive stages.
The acceleration evaluation is based on a single Nvidia V100 GPU in the MS COCO dataset.
As shown in Table \ref{tab:4}, we can find that when the stage number increases from 8 to 64, the performance improves prominently with slower inference speed. When it increases to 256, the generation performance reaches a plateau.
Please note that the model with 1024 stages is actually the autoregressive generation model with our dynamic scoring-based strategy.
Therefore, we set the default stage number to 64 in our experiments for competitive performance and faster inference speed. 

\begin{figure}[t]
  \centering
   \includegraphics[width=1\linewidth]{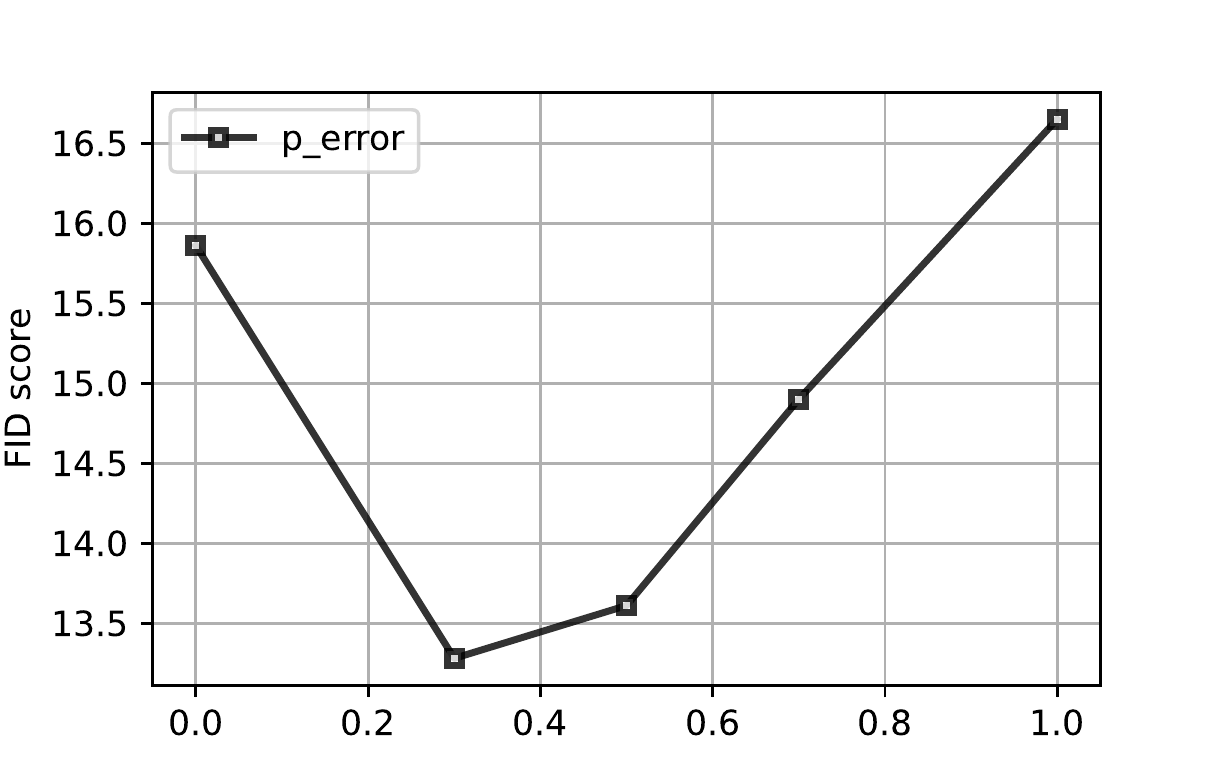}
   \caption{Effects of image token revision. }
\end{figure}

\begin{table*}[h]
	\begin{center}
	\setlength{\tabcolsep}{1.5mm}{
		\begin{tabular}{lcccccc}
		\hline
		&\multicolumn{3}{c}{\textbf{Image Realism}} &\multicolumn{3}{c}{\textbf{Image-Text Match}} \\ 
		&baseline wins&border&prog. wins &baseline wins &border &prog. wins \\ \hline 
		minDALL-E&34.3&24.2&41.5&33.0&31.2&35.8\\
		Random&31.2&29.7&39.2&33.7&27.8&38.5\\
		\hline
		\end{tabular}
		{\caption{Human evaluation results over 200 textual prompts and the corresponding generated images from the MS COCO test set. }
			\label{tab:human}}
			}
	\end{center}
\end{table*}

\begin{figure*}[t]
\centering
\includegraphics[width=2\columnwidth]{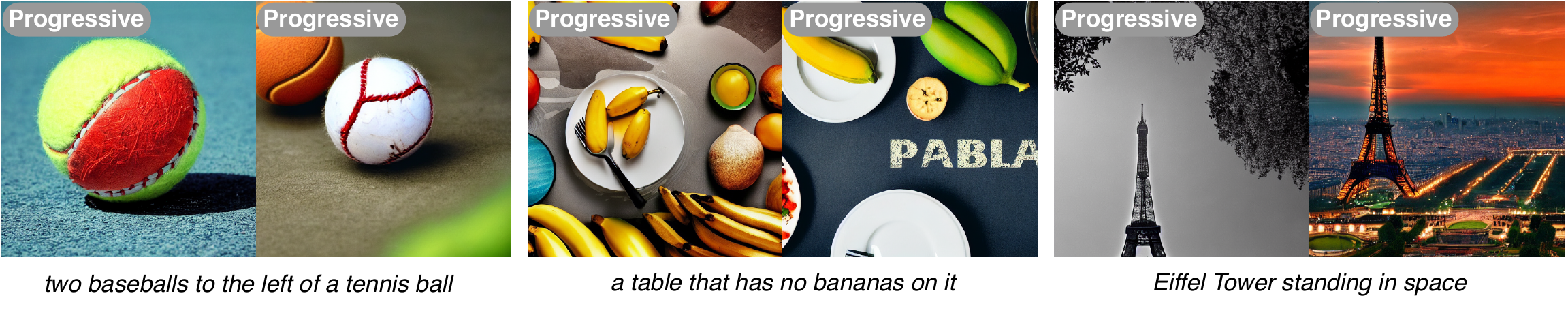} 
\caption{Images generated from progressive model showing errors in \emph{number counting and negative semantic understanding}, which motivates the future improvement.}
\label{fig:4}
\end{figure*}

\paragraph{Effect of Image Token Revision.}
We investigate the influence of error revision in this section, where $p_{error}$ is the probability of injecting pseudo incorrect image tokens to $T$ training series for each instance. 
From the experiment results, we can observe that: ($\textbf{i}$) without the error revision, i.e., $p_{error} = 0$, the FID score increases significantly, indicating that the error revision mechanism is effective for performance improvement. ($\textbf{ii}$)  As $p_{error}$ becomes larger, the performance becomes improved at first and deteriorated thereafter. We believe that too many pseudo errors make it hard to learn the correct text-to-image mapping. ($\textbf{iii}$) The model achieves best performance with $p_{error}=0.3$, which is set as default value in other experiments.


\subsection{Cases from Progressive Model}
\label{sec:4.3}
To demonstrate the capability of generating in-the-wild images, we provide some generated results in Figure \ref{fig:case} and more intuitive cases comparison can be found in Appendix \ref{app:case}. Though our base model is much smaller than previous works like Parti, we also achieve a promising performance with some delicate design. 
Compared with the AR methods which generate images from top-left to down-right, our method generates images in a global manner and supports error revision timely, resulting in much more high-quality and content-rich images. 
We also list some bad cases in Figure \ref{fig:4} to provide insights on how our approach may be improved. 
These cases show representative types of errors, \emph{i.e.}, negative semantic understanding, and spatial relations. 
Although our approach generates unfavorable images, it still generates related subjects.


\subsection{Human Evaluation}
\label{sec:4.4}

We follow the form in \cite{yu2022scaling} to conduct side-by-side human evaluations for minDALL-E, Random vs. our progressive model from image realism and image-text match aspects.
Please note that compared with minDALL-E, the model with random generation mode uses image token revision.
Detailed setting please refer to appendix \ref{app:human_eval}. The evaluation results are summarized in Table \ref{tab:human}. As we can see, our progressive model outperforms minDALL-E, which is the popular open-source autoregressive image generation model and holds a similar model parameter and training data size. When compared against the random mode with the same network architecture, our progressive model still shows superiority for optimized generation order.

\section{Related Works}

\paragraph{Autoregressive Image Synthesis.} 


Currently, autoregressive models \cite{radford2018improving,radford2019language} have shown promising results for text-to-image generation \cite{chen2020generative,esser2021taming,van2017neural,parmar2018image,razavi2019generating,van2016pixel}. Prior works including PixelRNN \cite{van2016pixel}, Image Transformer \cite{parmar2018image} and ImageGPT \cite{chen2020generative} factorize the conditional probability on an image over raw pixels. Due to the intolerable amount of computation for large images, modeling images in the low-dimension discrete latent space is introduced. VQ-VAE, VQ-GAN, and ImageBART \cite{esser2021imagebart} train an encoder to compress the image and fit the density of the hidden variables. It greatly improves the performance of image generation. More recent DALL-E \cite{ramesh2022hierarchical}, CogView \cite{ding2021cogview}, M6 \cite{lin2021m6}, ERINE-ViLG \cite{zhang2021ernie}, and Parti \cite{yu2022scaling} all utilize AR-based Transformer architectures in the latent space. Similarly, \cite{lee2022draft} consider global image information with refinement by random masking. With a powerful large transformer structure and massive text-image pairs, they greatly advance the quality of text-to-image generation yet still ignore the importance and order of image tokens.

\paragraph{Denoising Diffusion Probabilistic.}
Another related work for text-to-image generation is deep diffusion model, which is first proposed in \cite{sohl2015deep} and achieved strong results on audio \cite{kong2020diffwave,jeong2021diff}, image  \cite{dhariwal2021diffusion,ho2020denoising,ho2022cascaded,nichol2021improved,saharia2022palette}, video \cite{ho2022video} generation, and super super-resolution \cite{saharia2021image}. Discrete diffusion models are also first described in \cite{sohl2015deep}, and then applied to text generation \cite{hoogeboom2021argmax,austin2021structured}. D3PMs \cite{austin2021structured} 
introduce discrete diffusion to image generation. 
As directly estimating the density of raw image pixels can only generate low-resolution images, more recent works\cite{rombach2022high,gu2022vector,wang2022clip,lee2022progressive} resort to diffuse in the VQ-based latent space. 

\section{Conclusion}

In this paper, we propose that the generation order of image tokens are important for text-to-image generation.
To this end, we introduce a progressive model, which builds the image sequence in a coarse-to-fine manner according to variant scoring strategies. The resulting top-down hierarchy makes the generation process interpretable and enjoys a significant reduction over empirical time.
Moreover, we seamlessly integrate the component of image token revision into our progressive framework, which further improves the model performance.
Extensive experiments show that our progressive model can produce more perceptually appealing samples and better evaluation metrics than conventional autoregressive models.
More encouragingly, our model achieves much faster inference speed, and is looking forward to be applied to various practical applications.

{\small
\bibliographystyle{ieee_fullname}
\bibliography{egbib}
}

\clearpage

\appendix

\section{Experimental Settings}
\label{app:exp_setting}

\begin{figure*}[t]
\centering
\includegraphics[width=1.8\columnwidth]{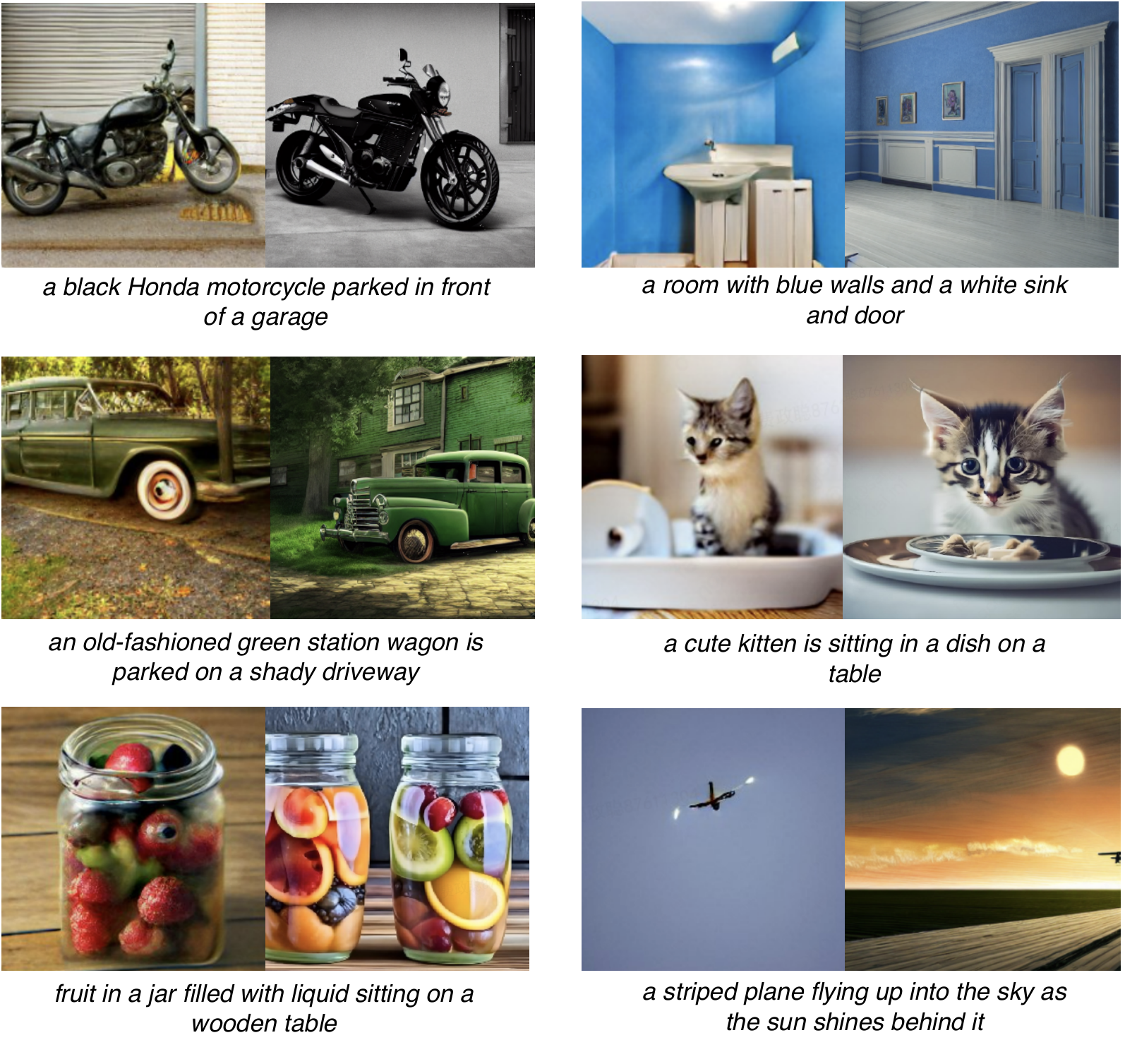} 
\caption{Image samples generated from minDALL-E in the left and our progressive model in the right, equipped with input text prompts from MS COCO dataset.}
\label{fig:app_case}
\end{figure*}

\paragraph{Datasets for Training and Evaluation.} 

We train on a combination of image-text datasets for progressive model training, including a filtered subset of LAION-400M and Conceptual Captions-3M. For all image inputs, we follow the VQ-GAN \cite{esser2021taming} input processing with weight trained on ImageNet to pre-extract the image token sequence.  
To demonstrate the capability of our proposed method for text-to-image synthesis, we conduct experiments on the MS COCO dataset \cite{lin2014microsoft}, which currently is a standard benchmark for text-to-image performance evaluation. MS COCO dataset contains 82k images for training and 40k images for testing. Each image in this dataset has five human-annotated text descriptions. In this paper, we conduct experiments consistent with Karpathy split \cite{karpathy2015deep}.

\paragraph{Implementation Details.}

For the image tokenizer, we follow the setting of original VQ-GAN \cite{esser2021taming}, which leverages the GAN loss to get a more realistic image. The codebook size is 163,84 with a dimension of 256, and a compression ratio of 16. That is, it converts $512 \times 512$ images into $32 \times 32$ tokens. We directly adopt the publicly available VQ-GAN model trained on the ImageNet dataset for all text-to-image synthesis experiments from  {https://github.com/CompVis/taming-transformers}.
We adopt a publicly available tokenizer of the base version of t5 \cite{raffel2020exploring} as a text encoder.
For the decoder of the text-to-image transformer, we set the stacked layer number to 24, the hidden dimension to 1280, the feed-forward dimension to 4096, and the head number to 20. An additional linear layer is appended at the last transformer layer to predict the state sequence. 
For error revision data construction, we select $p_{error}=0.3$ with a fixed 15\% replaced ratio of available tokens in the current sequence by default.
Besides, more advanced strategies for pseudo image token selection are left for future work. 
Both image and text encoders in our training process are frozen.
We use AdamW \cite{kingma2014adam} optimizer with $\beta_1$ = 0.9 and $\beta_2$ = 0.96. The model is trained for 120 epochs with the cosine learning rate schedule with the initial value of 1e-4.

\section{More Cases Analysis}
\label{app:case}

To illustrate the performance of the proposed progressive model more intuitively, we also compared it with the most popular VQ-AR based minDALL-E model. The generated images can be seen in Figure \ref{fig:app_case}, where the input text prompts are from the MS COCO dataset. We can observe that the results of the progressive model are more fine-grained,  more harmonious from a global perspective, and the semantics controls are more accurate.

\section{Human Evaluation} 
\label{app:human_eval}

We follow \cite{yu2022scaling} to conduct side-by-side human evaluations, in which well-educated human annotators are presented with two outputs for the same prompt and are asked to choose which image is a higher quality and more natural image (image realism) and which is a better match to the input prompt (image-text alignment). As for the Turing test, the model types are anonymized and randomly shuffled for each presentation to an annotator, and each pair is judged by three independent annotators. 
The results are summarized in Table \ref{tab:human}.
Finally, annotators have received reasonable remuneration for their labor. 
\end{document}